\documentclass[runningheads]{llncs}
\usepackage[T1]{fontenc}
\usepackage{graphicx}
\usepackage{multirow}

\begin{document}
\title{Classification in Histopathology: A unique deep embeddings extractor for multiple classification tasks}
\titlerunning{Efficient Classification of Histopathology Images with Deep Embeddings}

%


\author{Adrien Nivaggioli\inst{1} \and
Nicolas Pozin\inst{1} \and
Rémy Peyret\inst{1} \and
Stéphane Sockeel\inst{1} \and
Marie Sockeel\inst{1} \and
Nicolas Nerrienet\inst{1} \and
Marceau Clavel\inst{1} \and
Clara Simmat\inst{1} \and
Catherine Miquel \inst{2}
}
\authorrunning{A. Nivaggioli et al.}

%

\institute{Primaa, Paris, France  \\ \email{correspondence@primaalab.com} \and Hôpital Saint Louis, Paris, France}

\maketitle              
\begin{abstract}

In biomedical imaging, deep learning-based methods are state-of-the-art for every modality (virtual slides, MRI, etc.) In histopathology, these methods can be used to detect certain biomarkers or classify lesions. However, such techniques require large amounts of data to train high-performing models which can be intrinsically difficult to acquire, especially when it comes to scarce biomarkers.
To address this challenge, we use a single, pre-trained, deep embeddings extractor to convert images into deep features and train small, dedicated classification head on these embeddings for each classification task. This approach offers several benefits such as the ability to reuse a single pre-trained deep network for various tasks; reducing the amount of labeled data needed as classification heads have fewer parameters; and accelerating training time by up to 1000 times, which allows for much more tuning of the classification head.
In this work, we perform an extensive comparison of various open-source backbones and assess their fit to the target histological image domain. This is achieved using a novel method based on a proxy classification task. We demonstrate that thanks to this selection method, an optimal feature extractor can be selected for different tasks on the target domain. We also introduce a feature space augmentation strategy which proves to substantially improve the final metrics computed for the different tasks considered.
To demonstrate the benefit of such backbone selection and feature-space augmentation, our experiments are carried out on three separate classification tasks and show a clear improvement on each of them: microcalcifications classification (29.1\% F1-score increase), lymph nodes metastasis detection (12.5\% F1-score increase), mitosis classification (15.0\% F1-score increase).
We therefore show how an optimal unique feature extractor can be used in a multi-task setting, with improved performance, faster training and inference and optimized memory usage.

\end{abstract}

\newpage
\section{Introduction}
Digital pathology is rapidly gaining popularity as a reliable and efficient tool for medical diagnosis \cite{Dawson2022}. It paves the way for Artificial Intelligence (AI) based techniques usage as an aid to diagnosis, leading to significant improvements in accuracy and efficiency \cite{Zhou2021,Suganyadevi2021,Banerji2021,Wu2022}. However, despite recent advancements in AI and computer vision, their application in digital pathology faces several challenges. For instance, training a deep learning model requires a large volume of annotated data to avoid overfitting, which can be difficult to acquire. Indeed, annotating medical data is a time-consuming and expensive process, as it requires dedicated software and trained professionals. Deep learning algorithms also require significant computational resources, which can limit experimentation and parameter tuning. Furthermore, some biomarkers are quite rare (vascular emboli, perineural invasion,...), leading to small amounts of data, which makes it difficult to train accurate deep learning models.

To address these challenges, we use a pretrained deep learning backbone to convert images into embeddings. The embeddings are then used to train small dedicated neural networks. In this study, we focus on histopathology classification tasks, and refer to these networks as "classification heads". As a consequence, we alleviate the need for labelled data and make the training process shorter as only small specific networks need to be trained. Furthermore, we propose a way to compare and select the best backbone to use to extract embeddings, and we prove the importance of the augmentations in the feature space.

In recent years, a few papers have followed similar so-called Deep Feature Learning (DFL) approaches  \cite{Kumar2020}. However, most of them use an arbitrarily chosen backbone while other existing models may be more relevant to their usage \cite{Kumar2020}. Most authors also focus their analysis to a single use case while an efficient feature extractor may be used for multiple tasks \cite{Spanhol2017}.

In this paper, we first describe our backbone selection method and perform a large-scale benchmark of existing open-source networks to find the most relevant for histopathology images. Then, we detail various experiments on histopathology classification tasks and show that a DFL-based method, using the same backbone, outperforms traditional deep learning on every tested classification tasks. We discuss our results and show the importance of data augmentation in the feature space, as it greatly improves performances.

\subsubsection{Contributions}

\begin{itemize}
\item We propose a backbone selection method. It is applied on several open source models so as to select a backbone relevant to our tasks.
\item We prove the effectiveness of feature space data augmentations.
\item We successfully apply our Deep Feature Learning based methodology on three histopathology classification tasks.
\item  We prove the superiority of our approach compared to standard deep learning approaches in digital pathology.
\end{itemize}

\begin{figure}
\includegraphics[width=\textwidth]{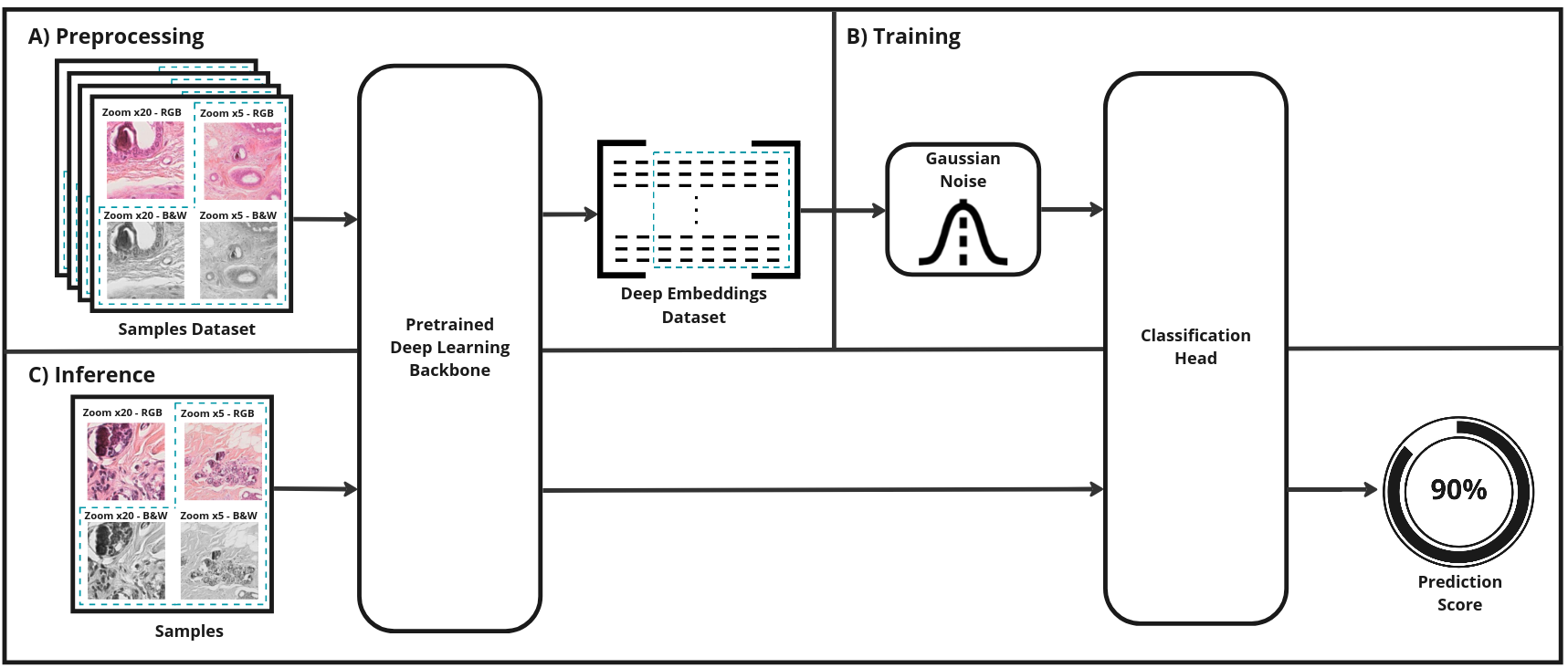}
\caption{Overview of our method. A) A sample is a RGB image at zoom x20. It is enriched with configurable variations such as: sibling patch at zoom x5, black and white translation of the patch...  Each element from the combination is converted into embeddings through a deep network. Resulting embeddings are concatenated to create a single deep embedding dataset. B) A small classification head is trained on the deep embedding dataset. C) At inference time, images are passed through the backbone and the classification head to get a prediction.} \label{fig1}
\end{figure}

\section{Pretrained Backbone as Universal Histopathological Feature Extractors}
\subsection{Concept}

Complex tasks may require the usage of deep networks to get satisfying results. Such cases where little data is available and big networks are required lead to overfitting. In digital pathology, the available datasets often are relatively small, as it is expensive and time consuming to build large annotated datasets that requires expert knowledge.

Thus, projecting images into a lower-dimensional feature space is a way to address these issues as we only train small classification networks on top of the backbone, reducing at the same time the need for labeled data, the risk of overfitting, and the training time.
 Moreover, we can still benefit from deep neural networks abstraction abilities, even if they are trained on external data.

The crux of our method is the assumption that images, which are high-dimensional (e.g., 256x256x3), can be projected into a smaller feature space by using deep learning task agnostic backbones without significant loss of information relevant for our classification tasks.

\subsection{Backbone Selection}

To ensure that deep embeddings carry sufficient information and to compare various backbones performances, we create an evaluation dataset composed of 35334 breast histopathology images at zoom x5 (1.76  $\mu$m  per  pixel) distributed amongst 23 imbalanced classes, which include both common tumor and benign lesions. This small dataset only required few annotated data, labeled in-house by expert pathologists. Moreover, it allows us to compare backbones on histopathological images, which are the types of images the following experiment are based on. The precise composition of this dataset is described in the supplementary materials, in Table~\ref{backbone_dataset}.

As a baseline, we train every layer of a ResNet50 \cite{he2016deep} on this dataset, starting with Tensorflow's \cite{tensorflowTensorFlow} ImageNet weights. We use a focal loss \cite{Lin2020} to fight the imbalance along with ADAM optimizer \cite{kingma2014method}, and various augmentations: scaling, shear, and flip. We achieve a maximum validation balanced accuracy of 0.6889.

For various open-source backbones, we train a classification head directly on normalized deep embedding extracted from our image dataset. Each head is composed of a single dense layer with a softmax activation, trained with a focal-loss, ADAM optimizer, and Gaussian Noise as augmentation. 

Table~\ref{tab2} provides an overview of our results. It is quite revealing in several ways. First, most pretrained backbones outperform the baseline model, which quickly overfits given the reduced size of the dataset.
At the moment, backbones pretrained on Medical Images perform poorly when compared to those pretrained of ImageNet-1K or ImageNet-22K. This confirms what recent works have found \cite{Tavolara2022,chen2022self}. One possible explanation for this is the fact that the best performing available models have been trained in an unsupervised manner for a long time on large GPU clusters \cite{caron2021emerging}, whereas those pretrained on Medical Images had access to more limited ressources and less diverse data \cite{chen2022self}. This might change in the future with the increasing amount of open-source medical datasets and research teams working with histopathological images.
Vision Transformers perform worse than regular convolution-based backbones. This is in line with findings of previous works, which did not find clear advantages of vision-based models in comparison with standard convolution-based ones \cite{CNNRules,Springenberg2022FromCT}.
Backbones trained in an unsupervised manner \cite{caron2021emerging,swav,rao2022hornet} perform better than their supervised counterparts. This is coherent with the rise of foundation models, which are taking over the leaderboards of many Image Processing datasets.

A visual comparison of the embeddings of the evulation dataset using the HistoSegNet \cite{Chan_2019_ICCV} and DINO ResNet-50 \cite{caron2021emerging} backbones is shown in the supplementary materials, in Figure~\ref{Umap}.

\begin{table}
\caption{Comparison of various pretrained backbones}\label{tab2}
\begin{tabular}{|l|c|c|c|c|c|}
\hline
\multicolumn{1}{|c|}{\textbf{Pretrained}} & \multicolumn{1}{c|}{\textbf{Pretrained on}}& \multicolumn{1}{c|}{ \textbf{Unsupervised} } & \multicolumn{1}{c|}{\textbf{Balanced} } & \multicolumn{1}{c|}{\textbf{Cohen Kappa} } & \multicolumn{1}{c|}{ \textbf{Weighted} }\\ 
      \multicolumn{1}{|c|}{\textbf{Backbone}} & \textbf{Medical Images} & \textbf{Training} & \textbf{Accuracy} & \textbf{Score} & \textbf{F1-Score} \\ \hline \hline
    HistoSegNet \cite{Chan_2019_ICCV} & yes & no & 0.5676 & 0.7942 & 0.8259 \\    \hline
    ResNet50 \cite{chen2022self} & yes & yes & 0.6519 & 0.8194 & 0.8498 \\    \hline
    ViT-S/16 \cite{chen2022self} & yes & yes & 0.6522 & 0.8444 & 0.8706 \\    \hline
    ViT-patch16-224 \cite{wu2020visual} & no & no &0.7098 & 0.8904 & 0.9090 \\ \hline
    DINO ViT-S/8 \cite{caron2021emerging} & no & yes &0.7287 & 0.8978 & 0.9149 \\    \hline
    Keras's ResNet50 & no & no & 0.7404 & 0.9029 & 0.9197 \\    \hline
    ConvNeXt-B \cite{convext} & no & no & 0.7686 & 0.9135 & 0.9292 \\ \hline
    HorNet-L \cite{rao2022hornet} & no & yes &0.7894 & 0.9276 & 0.9411 \\   \hline
    SwAV ResNet-50 \cite{swav} & no & yes & 0.8196 & 0.9290 & 0.9415 \\    \hline
    \textbf{DINO ResNet-50} \cite{caron2021emerging} & no & yes & \textbf{0.8236} & \textbf{0.9365} & \textbf{0.9475} \\    \hline
    \hline

    \textbf{ResNet-50} & \multirow{2}{*}{-} & \multirow{2}{*}{no} & \multirow{2}{*}{0.6582} & \multirow{2}{*}{0.7599} & \multirow{2}{*}{0.7995} \\
    \textbf{Fully Trained} &  &  &  &  & \\    \hline

\end{tabular}
\end{table}

\section{Histopathology Classification Experiments}

We apply the Deep Feature Learning method to various histopathology classification tasks: classification of Lymph Nodes Metastasis, Microcalcifications and Mitosis.

In histopathology, pathologists use information from various zoom levels to perform diagnosis as both context and cellular level patterns are of importance. Some AI based methods\cite{multiscale} replicate this by making prediction on multiple images at same position  but different zooms. 
Doing so with traditional deep learning requires specific architectures handling multiple input images, which increases the model dimensions and makes the method even more prone to overfitting. Also, increasing even more the input dimension by adding multiple input images facilitates overfitting. Using DFL, we can easily concatenate features extracted with a same backbone from samples at various zooms.
In the following experiments, we compare various approaches: Traditional Deep Learning at zoom x20, DFL at zoom x20, and DFL at both zoom x20 and x5.

Based on backbones comparisons illustrated on Table 1, we use the DINO ResNet-50 \cite{caron2021emerging} as backbone.
All trainings were performed on a single GeForce RTX 2080 Ti GPU, using a Focal Loss and an ADAM optimizer. Apart from the mitosis classification, traditional deep learning trainings used an EfficientNet architecture and performed online image augmentation (such as Color Jitter, Gaussian Noise, Hue/Saturation/Value random changes, random Flip and Rotations). In the DFL setting, only Gaussian noise augmentation is applied, directly on the extracted embeddings. The classification head is composed of a single dense hidden layer, followed by the classification layer. Figure~\ref{fig1} depicts an overview of our method.

\begin{table}[]
\caption{Dataset description}\label{dataset_description}
\begin{tabular}{|c|c|c|c|c|c|}
\hline
\textbf{Biomarker} & \textbf{Set} & \textbf{\#Slides} & \textbf{\begin{tabular}[c]{@{}c@{}}\#Patches\\ Positives/Negatives\end{tabular}} & \textbf{Origin} & \textbf{\begin{tabular}[c]{@{}c@{}}Sample \\ Composition\end{tabular}} \\ \hline \hline
\multirow{4}{*}{\begin{tabular}[c]{@{}c@{}}Lymph Nodes\\  Metastasis\end{tabular}} &   \multirow{2}{*}{Train}   & \multirow{2}{*}{500} & \multirow{2}{*}{720363 / 869493} &   \multirow{2}{*}{Camelyon17  \cite{Pinchaud2019Camelyon17C}}   & \multirow{4}{*}{\begin{tabular}[c]{@{}c@{}}Zooms x20 and x5\\ B\&W Images\end{tabular}} \\
 &  &  &  &  &  \\ \cline{2-5}
 & \multirow{2}{*}{Test} & \multirow{2}{*}{50} & \multirow{2}{*}{57334 / 173050} & \multirow{2}{*}{In-House} &  \\
 &  &  &  &  &  \\ \hline \hline
\multirow{4}{*}{Microcalcifications} & \multirow{2}{*}{Train} & \multirow{2}{*}{1642} & \multirow{2}{*}{2825 / 331842} & \multirow{4}{*}{\begin{tabular}[c]{@{}c@{}}In-House\\ 3 Centers\end{tabular}} & \multirow{4}{*}{\begin{tabular}[c]{@{}c@{}}Zooms x20 and x5\\ RGB and B\&W Images\end{tabular}} \\
 &  &  &  &  &  \\ \cline{2-4}
 & \multirow{2}{*}{Test} & \multirow{2}{*}{284} & \multirow{2}{*}{632 / 57421} &  &  \\
 &  &  &  &  &  \\ \hline \hline
\multirow{4}{*}{Mitosis} & \multirow{2}{*}{Train} & \multirow{2}{*}{180} & \multirow{2}{*}{2500 / 2500} & \multirow{2}{*}{\begin{tabular}[c]{@{}c@{}}MIDOG21 \cite{grandchallengeMIDOGChallenge} \\ In-House\end{tabular}} & \multirow{4}{*}{\begin{tabular}[c]{@{}c@{}}Zoom x20\\ RGB and B\&W Images\\ Padding and Scaling\end{tabular}} \\
 &  &  &  &  &  \\ \cline{2-5}
 & \multirow{2}{*}{Test} & \multirow{2}{*}{10} & \multirow{2}{*}{125 / 1108} & \multirow{2}{*}{In-House} &  \\
 &  &  &  &  &  \\ \hline
\end{tabular}
\end{table}

Test and train datasets used for every classification tasks are detailed in Table~\ref{dataset_description}. Note that samples composition for the DFL (considered zoom, color images on which embeddings are extracted...) may be different for each task. As explicited in Table~\ref{dataset_description}, lymph nodes and microcalcification tasks are performed on samples containing images at both zooms x20 and x5 as context matter for theses tasks. For the mitosis classification task, high resolution details is necessary, and only zoom x20 is used. In addition, it was found sufficient to extract embeddings from RGB images for the lymph nodes metastasis task while features extracted from a combination of RGB and black and white images were more suited for mitosis and microcalcififications classification.
Most images had a size of 256*256 pixels, apart from mitosis images that had a size of 50*50 pixels. To make them fit into our backbone, we used a concatenation of scaling and padding to 256*256 pixels.
Finally, some datasets used for our experiments are public while some were labeled on in-house data by expert pathologists following a standard process: annotation is performed by a junior pathologist and then reviewed by an expert.

Table~\ref{tab_biomarker} summarizes our results and proves the efficiency of the Deep Feature Learning methodology over traditional Deep Learning

\begin{table}[]
\caption{Biomarker Classification}\label{tab_biomarker}
\begin{tabular}{|c|c|c|c|c|c|}
\hline
\textbf{Biomarker} & \textbf{Method} &    \textbf{Precision}    &    \textbf{Recall}    &    \textbf{PR-AUC}    &    \textbf{Training time}    \\ \hline \hline
\multirow{3}{*}{\begin{tabular}[c]{@{}c@{}}Lymph Nodes\\ Metastasis\end{tabular}} & Deep Learning & 0.848 & 0.713 & 0.848 & 7 days \\ \cline{2-6} 
 & DFL - Single Zoom & 0.856 & 0.769 & 0.882 & 17 minutes \\ \cline{2-6} 
 &   \textbf{DFL - Multi Zoom}   & \textbf{0.927} & \textbf{0.822} & \textbf{0.932} & \textbf{30 minutes} \\ \hline \hline
  \multirow{3}{*}{Microcalcifications}   & Deep Learning & 0.604 & 0.590 & 0.584 & 20 hours \\ \cline{2-6} 
 & DFL - Single Zoom & 0.614 & 0.603 & 0.619 & 15 minutes \\ \cline{2-6} 
 & \textbf{DFL - Multi Zoom} & \textbf{0.785} & \textbf{0.757} & \textbf{0.739} & \textbf{20 minutes} \\ \hline \hline
\multirow{2}{*}{Mitosis} & Deep Learning & 0.546 & 0.509 & 0.507 & 5 hours \\ \cline{2-6} 
 & \textbf{DFL - Single Zoom} & \textbf{0.574} & \textbf{0.641} & \textbf{0.584} & \textbf{4 minutes} \\ \hline
\end{tabular}
\end{table}

\section{Discussion}
\subsection{Backbone Selection Validation}
We tested the backbone selection method on the microcalcification classification task. Deep embeddings were extracted using various backbones, and the same classification head was trained on those embeddings. Our results, presented in Table~\ref{tab_backbon_comp2}, are consistent with those shown in Table~\ref{tab2}, and support the use of DINO ResNet-50 \cite{caron2021emerging} as the backbone of choice for our experiments.

\begin{table}[]
\caption{Comparing backbones for microcalcifications classification}\label{tab_backbon_comp2}
\begin{tabular}{|c|c|c|c|}
\hline
\textbf{Backbone} &     \textbf{Precision}     &     \textbf{Recall}     &     \textbf{PR-AUC}     \\ \hline  \hline
    Keras's ResNet-50     & 0.478 & 0.419 & 0.379 \\ \hline
HorNet-L \cite{rao2022hornet} & 0.593 & 0.426 & 0.435 \\ \hline
SWaV ResNet-50 \cite{swav} & 0.610 & 0.435 & 0.460 \\ \hline
\textbf{DINO ResNet-50} \cite{caron2021emerging}  & \textbf{0.714} & \textbf{0.541} & \textbf{0.607} \\ \hline
\end{tabular}
\end{table}

\subsection{Deep Learning versus Deep Feature Learning}
Results presented in Table~\ref{tab_biomarker} support our claims:
\begin{itemize}
  \item Trainings are much shorter. Instead of hours (or even sometimes days), they only take a few minutes. This allows for more parameters finetuning, and overall experimentation on the trainings and the dataset used.
  \item The results are better overall when using the same zoom as in the Deep Learning setting. This is due to the increased resistance to overfitting of the DFL method, as the Deep Learning setting inevitably overfitted even with extensive image augmentation,.
  \item In a DFL setting, using multiple zooms is trivial as it simply entails extracting several sets of embeddings during the preprocessing step.
\end{itemize}

Given a large enough dataset, traditional deep learning might outperform our DFL-based approach, which is dependent on the quality of the extracted embeddings. But, for our usecases and up to a couple million samples in the training dataset, our method still outperformed traditional deep learning.

\begin{table}[]
\caption{Augmentation Influence}\label{tab_augmentation}
\begin{tabular}{|c|c|c|c|c|c|}
\hline 
Biomarker                               & Method              & Gaussian Noise & Precision      & Recall         & PR-AUC         \\ \hline \hline
\multirow{4}{*}{Lymph Nodes Metastasis} & Single Zoom         & No             & 0.843          & 0.729          & 0.863          \\ \cline{2-6} 
                                        & Single Zoom         & Yes            & 0.856          & 0.769          & 0.882          \\ \cline{2-6} 
                                        & Multi Zoom          & No             & 0.919          & 0.795          & 0.915          \\ \cline{2-6} 
                                        & \textbf{Multi Zoom} & \textbf{Yes}   & \textbf{0.927} & \textbf{0.822} & \textbf{0.932} \\ \hline \hline
\multirow{4}{*}{Microcalcifications}    & Single Zoom         & No             & 0.596          & 0.562          & 0.574          \\ \cline{2-6} 
                                        & Single Zoom         & Yes            & 0.614          & 0.603          & 0.619          \\ \cline{2-6} 
                                        & Multi Zoom          & No             & 0.701          & 0.619          & 0.678          \\ \cline{2-6} 
                                        & \textbf{Multi Zoom} & \textbf{Yes}   & \textbf{0.785} & \textbf{0.757} & \textbf{0.739} \\ \hline
\end{tabular}
\end{table}

\subsection{On Augmentations}
The main drawback of our method is the difficulty of using image augmentation. Because the image dataset is first preprocessed and transformed into a deep embeddings dataset, images cannot be augmented on the fly as usually done in traditionnal deep learning methods. 

A partial answer would be to augment the image dataset offline, before converting it to embeddings, but the resulting set would still be limited.
Another way is to perform augmentation on the fly directly in the feature space. Here, we apply gaussian noise directly on the deep embeddings. This is paramount, as the classification head otherwise overfits in a couple of epochs. Table~\ref{tab_augmentation} shows the importance of this noise, as it persistently improves performance.

We believe that on the fly augmentation in the feature space is key to push performance further. In the future, we will explore more refined ways than simple homogeneous gaussian noise: various statistical perturbations, various augmentations for the different components of the feature vector, sampling in the feature space to simulate given image augmentations (e.g color augmentation) or image synthesis...

\subsection{On Concatenating Deep Embeddings}
We find that extracting and concatenating multiple embeddings for each image often led to better results. When training classification heads, we tried several combinations of deep embeddings by varying the image zoom (x5 or x20) and the color image color space (RGB or Black and White).

An exploration path would be to extract deep embeddings with multiple backbones. By using multiple models trained on different datasets, with various training methodology and architectures, we may extract more representative deep embeddings \cite{gontijo-lopes2022no}. We could also further reduce the feature space dimension through dimension reduction techniques such as PCA, LDA, or Isomap, this would further speed up training time and reduce classification head sizes.

\section{Conclusion}
In this paper, we proved that a Deep Feature Learning based approach outperforms traditional deep learning on multiple histopathology classification tasks. Faster trains, lighter networks, and smaller datasets allows for more experimentation with limited computational ressources.
We proposed a backbone selection method that we used to compare various open-source backbones so as to find the most relevant for histopathology classification tasks. Our method eases the use of multiple images as inputs, which greatly improves results. The use of gaussian noise as an on-the-fly augmentation also proves valuable, and increases performances considerably.

Our DFL-based method is modular and, as discussed in the previous section, many avenues are still left for us to explore. Our method could in particular be adapted to detection and segmentation tasks, which would notably broaden its use.

\newpage

%
%
%
\bibliographystyle{splncs04}
\bibliography{biblio}

\newpage

\section{Supplementary Materials}

\begin{table}[]
\caption{Backbone Evaluation Dataset Composition}\label{backbone_dataset}
\begin{tabular}{|l|c|c|}
\hline
\textbf{Pathology} & \textbf{Training Dataset} & \textbf{Validation Dataset} \\
\hline \hline
Invasive Lobular Carcinoma & 9271 & 3091 \\ \hline
Fibroadenomas & 3797 & 1266 \\ \hline
Invasive Ductal Carcinoma & 4134 & 1378 \\ \hline
Healthy Tissue Glands/Lobules & 2747 & 916 \\ \hline
Simple Adenosis & 1995 & 665 \\ \hline
Ductal carcinoma in situ & 1403 & 468 \\ \hline
Cyst & 661 & 220 \\ \hline
Not Interpretable & 422 & 141 \\ \hline
Lymph Node & 397 & 132 \\ \hline
Lobular Carcinoma in situ & 301 & 100 \\ \hline
Microcalcifications & 263 & 88 \\ \hline
Healthy Fibrous Tissue & 222 & 74 \\ \hline
Apocrine Metaplasia & 164 & 55 \\ \hline
Columnar Cell Metaplasia & 156 & 52 \\ \hline
Inflammatory Lesion & 103 & 35 \\ \hline
Healthy Adipose Tissue & 98 & 33 \\ \hline
Sclerosing Adenosis & 94 & 31 \\ \hline
Papilloma & 85 & 28 \\ \hline
Simple Ductal Hyperplasia & 58 & 19 \\ \hline
Atipycal Ductal Hyperplasia & 41 & 13 \\ \hline
Blood Vessel & 38 & 12 \\ \hline
Scar & 27 & 9 \\ \hline
Necrosis & 23 & 8 \\ 
\hline
\end{tabular}
\end{table}

\begin{figure}
\includegraphics[width=\textwidth]{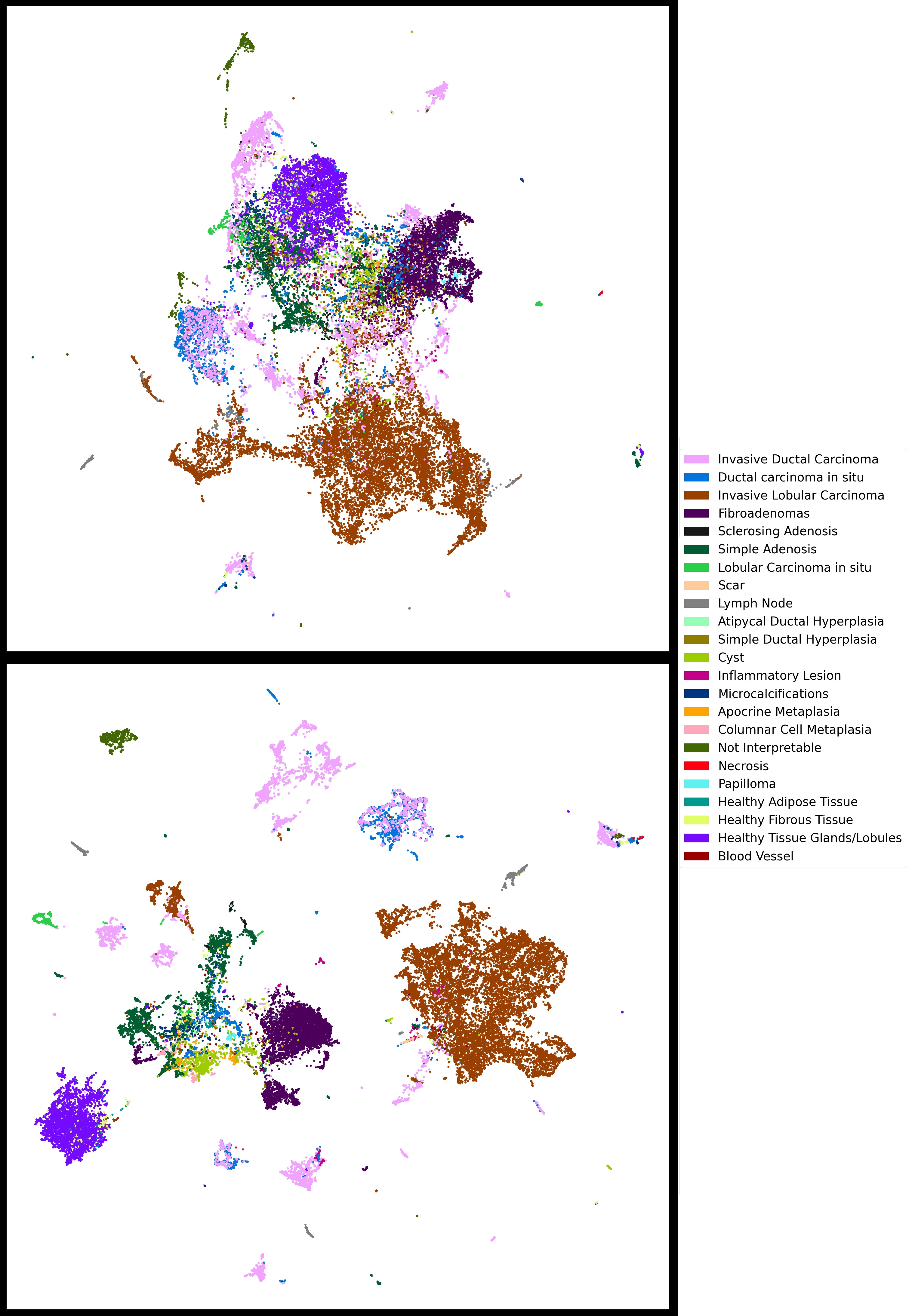}
\caption{Projection in 2 dimensions of the evaluation dataset embeddings using the HistoSegNet (top) and DINO ResNet-50 (bottom) backbones (Best viewed in color). The projection was performed using the UMAP algorithm, which had no information regarding the class of each image. The separation of the classes is much clearer using the DINO ResNet-50 backbone, and some classes already lie in their own, distinct clusters, which is consistent with the performance observed on the multiclass classification task used in the backbone selection method. This illustrates the quality of the feature extractor for the histopathology domain.} \label{Umap}
\end{figure}

\end{document}